\documentclass[10pt,twocolumn,letterpaper]{article}

\usepackage[pagenumbers]{cvpr} % To force page numbers, e.g. for an arXiv version

\usepackage[accsupp]{axessibility} % Improves PDF readability for those with disabilities.
\usepackage{graphicx}
\usepackage{amsmath}
\usepackage{amssymb}
\usepackage{booktabs}
\usepackage{multirow}

\def\ie{\emph{i.e.}}
\def\eg{\emph{e.g.}}

\usepackage[pagebackref,breaklinks,colorlinks,bookmarks=false]{hyperref}

\usepackage[capitalize]{cleveref}
\crefname{section}{Sec.}{Secs.}
\Crefname{section}{Section}{Sections}
\Crefname{table}{Table}{Tables}
\crefname{table}{Tab.}{Tabs.}

\def\cvprPaperID{5386} % *** Enter the CVPR Paper ID here
\def\confName{CVPR}
\def\confYear{2022}

\begin{document}

%%%%%%%%% TITLE - PLEASE UPDATE
\title{End-to-End Compressed Video Representation Learning for \\ Generic Event Boundary Detection}

\author{Congcong Li$^1$\thanks{This work was done during internships at ByteDance Inc.}, Xinyao	Wang$^2$, Longyin Wen$^2$, Dexiang Hong$^{1\ast}$, Tiejian Luo$^1$, Libo Zhang$^3$\thanks{Corresponding author (libo@iscas.ac.cn)}\\
{$^1$University of Chinese Academy of Sciences, Beijing, China } \\
{$^2$ByteDance Inc., Mountain View, USA}\\
{$^3$Institute of Software Chinese Academy of Sciences, Beijing, China}\\
{\tt \small licongcong18@mails.ucas.edu.cn, \{xinyao.wang,longyin.wen\}@bytedance.com}\\
{\tt \small hongdexiang19@mails.ucas.edu.cn, tjluo@ucas.ac.cn, libo@iscas.ac.cn}
}
\maketitle
\vspace*{-0.05in}

%%%%%%%%% ABSTRACThttps://www.overleaf.com/project/611cfa214ccfca8d1da3ead1
\begin{abstract}
Generic event boundary detection aims to localize the generic, taxonomy-free event boundaries that segment videos into chunks. Existing methods typically require video frames to be decoded before feeding into the network, which demands considerable computational power and storage space. To that end, we propose a new end-to-end compressed video representation learning for event boundary detection that leverages the rich information in the compressed domain, \ie, RGB, motion vectors, residuals, and the internal group of pictures (GOP) structure, without fully decoding the video. Specifically, we first use the ConvNets to extract features of the I-frames in the GOPs. After that, a light-weight spatial-channel compressed encoder is designed to compute the feature representations of the P-frames based on the motion vectors, residuals and representations of their dependent I-frames. A temporal contrastive module is proposed to determine the event boundaries of video sequences. To remedy the ambiguities of annotations and speed up the training process, we use the Gaussian kernel to preprocess the ground-truth event boundaries. Extensive experiments conducted on the Kinetics-GEBD dataset demonstrate that the proposed method achieves comparable results to the state-of-the-art methods with $4.5\times$ faster running speed.
\end{abstract}

\begin{figure}[t]
  \centering
   \includegraphics[width=1.0\linewidth]{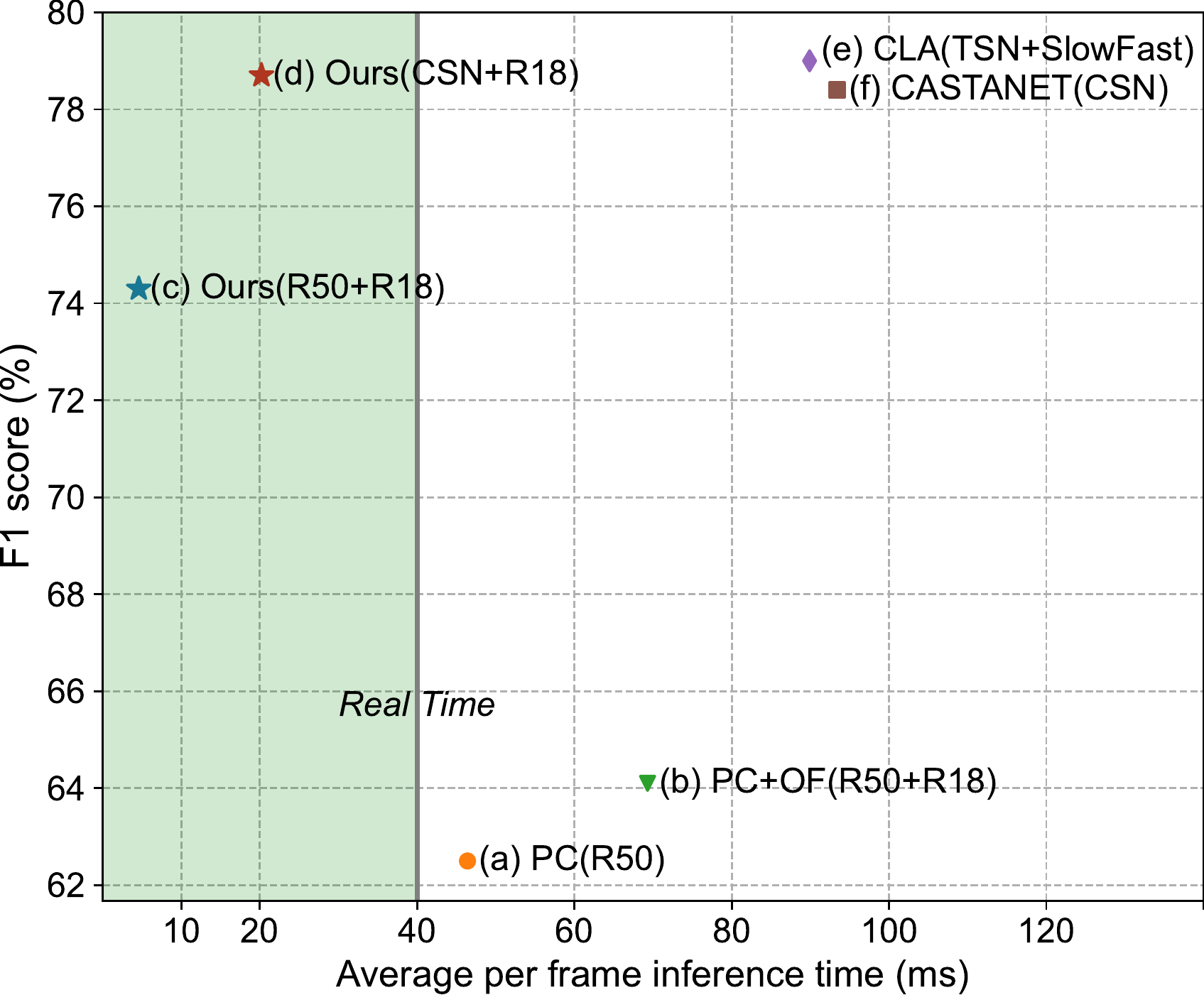}
   \caption{The inference time \textit{vs.} F1 score of different methods on the Kinetics-GEBD dataset \cite{DBLP:journals/corr/GEBD}. \textbf{(a)} The previous method \cite{DBLP:journals/corr/GEBD} PC is relatively fast with inferior results. \textbf{(b)} After integrating the optical flow (OF) module, the accuracy is improved with much slower running speed. \textbf{(c, d)} Our method achieves competitive F1 score with extremely fast running speed by directly leveraging motion vectors and residuals in the compressed domain. \textbf{(e, f)} CLA \cite{DBLP:journals/corr/abs-2106-11549} and CASTANET \cite{DBLP:journals/corr/abs-2107-00239} take fully decoded RGB frames as input, which are much slower than the methods conducted in compressed domain. The green region indicates the methods run in real-time.}
   \label{fig:onecol}
   \vspace*{-0.2in}
\end{figure}

%%%%%%% BODY TEXT
\vspace*{-0.1in}
\section{Introduction}
\vspace*{-0.1in}
\label{sec:intro}
Video traffic will account for 82\% percent of all internet traffic by 2022, up from 75\% in 2017 \cite{forecast2019cisco}. Understanding video content using AI technology is an active area of research in recent years. However, it is still a challenging task due to the complex temporal evolution in the enormous size of raw video streams with high temporal redundancy. 

Video understanding is one of the most fundamental problems in computer vision, which includes video tagging, action recognition, and video boundary detection, etc. In contrast to static images, videos provide rich information involving temporal consistency in consecutive frames which can be additionally utilized. Currently, the two-stream network \cite{DBLP:conf/nips/Two-Stream-Conv,DBLP:conf/cvpr/FeichtenhoferPZ16,DBLP:conf/iccv/Feichtenhofer0M19} and 3D convolutional network \cite{DBLP:conf/eccv/TaylorFLB10,DBLP:journals/pami/JiXYY13,DBLP:conf/iccv/TranBFTP15,DBLP:journals/pami/VarolLS18} are two popular network architectures in the video understanding field. The two-stream network incorporates both the decoded RGB video frames and optical flow to exploit temporal information. However, extracting optical flow is very slow, which dominates the overall pre-processing time in the video understanding tasks. 3D convolutional network is another choice to model temporal information using the spatio-temporal filters. The drawbacks of 3D convolutional network is the massive parameters contained in 3D convolution operations, which slows down the inference speed. Besides the aforementioned methods, the new trend in video understanding is using the transformers, including \cite{DBLP:conf/iclr/DosovitskiyB0WZ21,DBLP:journals/corr/abs-2103-15691,DBLP:journals/corr/abs-2106-13230,DBLP:journals/corr/abs-2104-11227,DBLP:conf/mm/ZhangHN21}, achieving competitive results.

In recent years, several methods \cite{DBLP:conf/cvpr/ZhangWW0W16,DBLP:conf/cvpr/coviar,DBLP:conf/cvpr/dmc-net,DBLP:conf/iccv/WangGLD19,DBLP:conf/iclr/YuLKS21,DBLP:conf/cvpr/HuangLWPXJ21} demonstrate the advantages of directly taking videos in compressed domain as input for video understanding. These methods use motion vectors and residuals in the compressed representation that developed for storage and transmission of videos rather than operating on the decoded RGB frames, which run in two orders of magnitude faster than the methods using optical flow while achieving competitive results \cite{DBLP:conf/cvpr/dmc-net}. Specifically, these methods use the almost compute-free motion vectors and residuals encoded in P-frames as an alternative to the compute-intensive optical flow. For example, CoViAR \cite{DBLP:conf/cvpr/coviar} directly feeds motion vectors and residuals into 2D CNNs for action recognition, and DMC-Net \cite{DBLP:conf/cvpr/dmc-net} improves the CoViAR method by reconstructing the optical flow based on motion vectors and residuals. Although the aforementioned method achieves promising results, they are still far from satisfactory, which lack effective fusion strategies between different modalities, such as decoded I-frames, motion vectors, and residuals. 

%In this paper, we focus on the generic event boundary detection task (GEBD) \cite{DBLP:journals/corr/GEBD} in compressed domain, which aims to localize the moments where humans naturally perceive taxonomy-free event boundaries that break a longer event into shorter temporal segments. The GEBD task is a challenging task since the event boundaries are taxonomy-free and the number of boundaries varies in different videos. The previous attempt \cite{DBLP:journals/corr/GEBD} formulate it as a classification task by considering the context information of the candidate boundaries. However, it neglects the temporal relations between consecutive frames. To that end, we design an end-to-end trained network to exploit the discriminative features for GEBD in the compressed domain, \ie, MPEG-4. Specifically, most modern codecs split a video into several group of pictures (GOP), where each GOP is formed by one I-frames and $T$ P-frames. To solve difficulty arised from the long chain of dependency of the P-frames, inspired by \cite{DBLP:conf/cvpr/coviar}, we use the back-tracing technique to compute the accumulated motion vectors and residuals in linear time. In this way, the consecutive P-frames in each GOP are only depending on the reference I-frame, which can be processed in parallel. 
In this paper, we focus on the generic event boundary detection (GEBD \cite{DBLP:journals/corr/GEBD}) task that aims to localize the moments where humans naturally perceive taxonomy-free event boundaries that segment a longer event into shorter temporal segments. The ability to divide a long form video into small meaningful clips makes this task demanding for several downstream video understanding tasks and industry applications that requires high accuracy and low latency. The previous attempt \cite{DBLP:journals/corr/GEBD} formulate it as a classification task by considering the context information of the candidate boundaries. However, it neglects the temporal relations between consecutive frames and operates inefficiently during feature extraction stage. Inspired by \cite{DBLP:conf/cvpr/ZhangWW0W16,DBLP:conf/cvpr/coviar,DBLP:conf/cvpr/dmc-net,DBLP:conf/iccv/WangGLD19,DBLP:conf/iclr/YuLKS21,DBLP:conf/cvpr/HuangLWPXJ21}, we design an end-to-end trained network to exploit the discriminative features for GEBD in compressed domain, \ie, MPEG-4, which is able to save decoding cost and improve feature extraction efficiency. Specifically, most modern codecs split a video into several group of pictures (GOP), where each GOP is formed by one I-frames and $T$ P-frames. To solve difficulty arised from the long chain of dependency of the P-frames, inspired by \cite{DBLP:conf/cvpr/coviar}, we use the back-tracing technique to compute the accumulated motion vectors and residuals in linear time. In this way, the consecutive P-frames in each GOP are only depending on the reference I-frame, which can be processed in parallel.

In contrast to the I-frame, it is difficult to learn the discriminative features of the P-frames. Refining the features of the reference I-frame based on the motion vectors and residuals becomes an intuitive option. Motion vectors and residuals provide information to reconstruct P-frames by referring the dependent I-frames. In addition to that, they also provide motion information that obtained from the video encoding process. To that end, we design a light-weight spatial-channel compressed encoder to refine the features of the reference I-frame with the guidance of the motion vectors and residuals. In this way, the features of P-frames and I-frames are converted to the same feature space, which benefits the subsequent processing. After that, a temporal contrastive module is proposed to capture the context information in temporal domain to predict the event boundaries of videos. Notably, our temporal contrastive module imitates humans, \ie, look back and forth around the candidate frames to determine event boundaries, by comparing the extracted features before and after the candidate frames. In addition, to remedy the ambiguities of annotations and speed up the training process, we use the Gaussian kernel to preprocess the ground-truth event boundaries instead of using the ``hard lables'' of boundaries. Extensive experiments conducted on the Kinetics-GEBD dataset to demonstrate the effectiveness of the proposed method. Specifically, the proposed method achieves comparable results to the state-of-the-art method at the CVPR'21 LOVEU Challenge \cite{DBLP:journals/corr/abs-2106-11549} with $4.5\times$ faster running speed, see Figure \ref{fig:onecol}.

The main contributions of this paper are listed as follows. (1) We propose an end-to-end compressed video representation learning method to solve the challenging GEBD task. (2) We design the spatial-channel compressed encoder to project the features of reference I-frame with the guidance of motion vectors and residuals to compute the features of P-frames with low cost. (3) A temporal contrastive module is proposed to determine the event boundaries of videos by exploiting the context information in temporal domain. (4) The proposed method achieves comparable results to the state-of-the-art methods at the CVPR'21 LOVEU Challenge \cite{DBLP:journals/corr/abs-2106-11549} with $4.5\times$ faster running speed, demonstrating its effectiveness.

\begin{figure*}[t]
\centering
\includegraphics[width=0.98\textwidth]{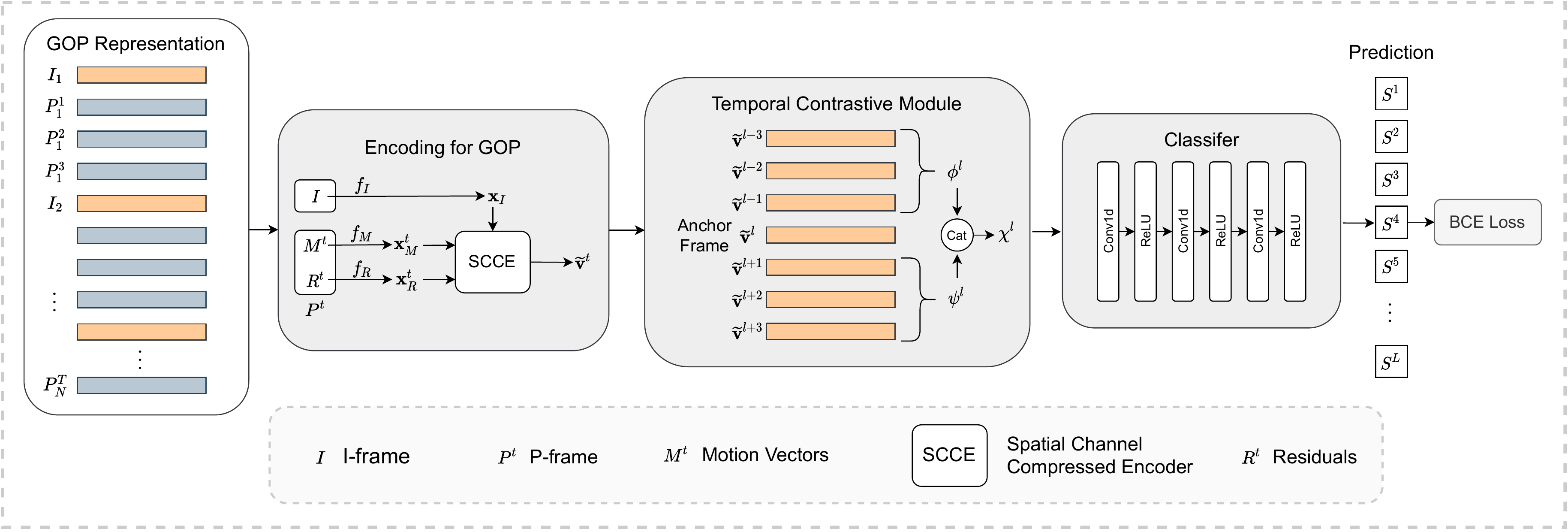}
\caption{The architecture of the proposed method. The spatial-channel compressed encoder (SCCE) is designed to obtain the refined P-frame representation $\widetilde{\bf v}^t$ based on reference I-frame feature ${\bf x}_I$, motion vectors $M^t$ and residuals $R^t$. This module regards each GOP as a process unit, which is efficient and can be paralleled in a large batch size. Then we use temporal contrastive module to capture temporal dependence explicitly based on unified representation $\widetilde{\bf v}^t$, which provides strong cues for boundary detection. After that, a simple classifier is used to make final predictions trained with the Gaussian smoothed soft labels.}
\label{fig:overall_architecture}
\vspace*{-0.1in}
\end{figure*}

%-------------------------------------------------------------------------
\vspace*{-0.1in}
\section{Related Work}
\vspace*{-0.1in}
\label{sec:related_work}
{\noindent \textbf{Video recognition.}} Over the last decade, video recognition has achieved great progress thanks to the emergence of deep learning. Early methods \cite{DBLP:conf/cvpr/WangKSL11,DBLP:conf/iccv/WangS13a,DBLP:conf/eccv/PengZQP14,DBLP:conf/cvpr/LanLLHR15} use hand-crafted features for video recognition. After the arriving of deep learning, the video recognition field is quickly dominated by the CNN-based methods, such as the two-stream network and 3D convolutional network. The two-stream network based methods \cite{DBLP:conf/nips/Two-Stream-Conv,DBLP:conf/cvpr/FeichtenhoferPZ16,DBLP:conf/iccv/Feichtenhofer0M19} use additional temporal stream to learn motion information and design various fusion strategies to combine the information from the image stream and temporal stream, achieving superior results. The optical flow is generally used to describe the motion information, which is computationally expensive. Some other methods \cite{DBLP:conf/eccv/TaylorFLB10,DBLP:journals/pami/JiXYY13,DBLP:conf/iccv/TranBFTP15,DBLP:journals/pami/VarolLS18} attempt to use the 3D convolutional network with the spatio-temporal filters to integrate temporal information. However, these methods are hard to optimize and require large-scale datasets in the training phase. The new trend in recent years is the introduction of transformers \cite{DBLP:conf/iclr/DosovitskiyB0WZ21,DBLP:journals/corr/abs-2103-15691,DBLP:journals/corr/abs-2106-13230,DBLP:journals/corr/abs-2104-11227,DBLP:conf/mm/ZhangHN21}, which achieving promising results on various datasets in video understanding.

Meanwhile, some recent methods attempt to directly take the raw compressed videos as input for different tasks in the video understanding field, such as action recognition \cite{DBLP:conf/cvpr/ZhangWW0W16,DBLP:conf/cvpr/coviar,DBLP:conf/cvpr/dmc-net,DBLP:conf/iclr/YuLKS21,DBLP:conf/cvpr/HuangLWPXJ21}, object detection \cite{DBLP:conf/iccv/WangGLD19}, and video segmentation \cite{DBLP:journals/corr/abs-2003-13260}. The aforementioned methods use motion vectors and residuals directly obtained from the compressed videos as the alternatives to optical flow and achieve comparable results in terms of both the speed and accuracy. 

{\noindent \textbf{Generic event boundary detection.}} Generic event boundary detection (GEBD) \cite{DBLP:journals/corr/GEBD} aims to localize the moments where humans naturally perceive taxonomy-free event boundaries that break a longer event into shorter temporal segments. The previous method \cite{DBLP:journals/corr/GEBD} takes $5$ video frames before and after the candidate boundaries as input, and separately determines whether each candidate is the event boundary or not. Kang \etal \cite{DBLP:journals/corr/abs-2106-11549} propose to use the temporal self-similarity matrix (TSM) as the intermediate representation and use the popular contrastive learning method to exploit the discriminative features for better performance. Hong \etal \cite{DBLP:journals/corr/abs-2107-00239} use the cascade classification heads and dynamic sampling strategy to boost both recall and precision. Meanwhile, Rai \etal \cite{DBLP:journals/corr/abs-2106-10090} attempt to learn the spatiotemporal features using a two stream inflated 3D convolutions architecture. To the best of our knowledge, there do not exist any prior work focuses on the GEBD task in the compressed domain.

{\noindent \textbf{Attention mechanism.}} To learn more discriminative features, numerous methods have been proposed, which mainly focus on enhancing the feature representations using the attention mechanisms on the spatial or(and) channel dimensions. SENet \cite{DBLP:conf/cvpr/SENet} develops the ``Squeeze-and-Excitation'' (SE) block that adaptively recalibrates channel-wise feature responses by explicitly modelling interdependencies between channels. Non-local network \cite{DBLP:conf/cvpr/non-local} capture long-range dependencies by computing the response at a position as a weighted sum of the features at all positions in the input feature maps. SKNet \cite{DBLP:conf/cvpr/SKNet} proposes to adaptively adjust the receptive field size of the input feature map by fusing multiple feature maps of different kernel sizes with softmax attention in a weighted manner. CBAM \cite{DBLP:conf/eccv/CBAM} sequentially infers attention maps along both channel and spatial dimensions, and then uses the attention maps to recalibrate the origin input feature. In contrast to the aforementioned methods, we attempt to refine the P-frame feature with the guidance of motion vectors and residuals by considering both spatial and channel dimensions of the features of I-frame, which fully leverages the information of the decoded reference I-frame to enrich the features of P-frames.

%------------------------------------------------------------------------
\vspace*{-0.1in}
\section{Method}
\vspace*{-0.1in}
\label{sec:method}
The existing method \cite{DBLP:journals/corr/GEBD} formulates the GEBD task as binary classification, which predicts the boundary labels of each frame by considering the temporal contextual information. That is, the preceding and succeeding frames of each video frame are feed into a neural network to detect the boundaries. It is inefficient due to the duplicated computation is conducted of consecutive frames. To remedy this, we propose an end-to-end compressed video representation method for GEBD, which regards each video clip as a whole. Specifically, we use MPEG-4 encoded videos as our input. Each video clip ${\cal V}$ is formed by $N$ groups of pictures (GOPs), and each GOP contains one I-frame and $T$ P-frames, \ie,
\begin{equation}
\begin{array}{ll}
    {\cal V}=\big\{I_i, P_i^1, P_i^2, \cdots, P_i^T\big\}_{i=1}^N,
\end{array}
    \label{equ:video_representation}
\end{equation}
where $I_i \in \mathbb{R}^{3 \times {\cal H} \times {\cal W}}$ denotes the reference I-frame and $P_i^t$ denotes the $t$-th P-frame of the $i$-th GOP, and ${\cal H}$ and ${\cal W}$ are the height and width of the video frame. For simplicity, we assume that there exists the same number of P-frames in all GOPs. The P-frame $P_i^t$ in the $i$-th GOP is formed by the motion vector $M_i^t \in \mathbb{R}^{2 \times {\cal H} \times {\cal W}}$ and residual $R_i^t \in \mathbb{R}^{3 \times {\cal H} \times {\cal W}}$, which can be obtained nearly cost-free from the compressed video stream. Notably, the motion vectors and residuals alone do not contain the full information of a P-frame. The P-frame depends on the reference I-frame or other P-frames, making it difficult to learn the discriminative feature representations for P-frames. Following \cite{DBLP:conf/cvpr/coviar}, we trace all motion vectors back until to the reference I-frame and accumulate the residual on the way to decouple the dependencies between the consecutive P-frames. In this way, each P-frame only depends on the reference I-frame rather than other P-frames. After that, we build our model based on the back-traced motion vectors and residuals and regard each GOP as a process unit. The overall network architecture is presented in Figure \ref{fig:overall_architecture}. As shown in Figure \ref{fig:overall_architecture}, the GOP is first encoded by the designed spatial-channel compressed encoder (SCCE) to generate the unified video representation. After that, a temporal contrastive module is used to exploit the temporal context information to get the discriminative feature representations. Finally, a classifier is used to generate the accurate event boundaries. 

\subsection{Spatial-Channel Compressed Encoder}
\label{sec:deformable_updating}
Motion, uncovered regions, and lighting variations frequently happen in video sequences. Modern codecs use macroblock as the basic unit for motion compensated prediction in a number of mainstream visual coding standards such as MPEG-4, H.263, and H.264. Motion vectors record the moving direction of each macroblock with respective to its reference frame(s), describing the motion patterns of videos, which is important for the GEBD task. The residuals can be regraded as the compensations of the motion information, which contains the boundary information of moving objects and plays a crucial role to identify the important regions in the I-frame. Thus, we propose to apply the attention mechanism to different regions of I-frame with the guidance of motion vectors to enrich the features by considering both channel and spatial dimensions. For simplicity, we omit the index $i$ of the GOP in the following sections.

Firstly, we use the convolutional neural network taking the decoded RGB image as input to extract the feature representation ${\bf x}_{I}$ of the I-frame $I$, \ie, ${\bf x}_{I} = f_I(I)$, where ${\bf x}_I \in \mathbb{R}^{C \times H \times W}$ is the features of the I-frame $I$, and $C$, $H$ and $W$ are the channel, height and width of the features ${\bf x}_I$, respectively. $f_I(\cdot)$ denotes the model used to extract features for I-frame, which is pretrained on the large-scale datasets (\eg, ResNet50 pretrained on ImageNet). Meanwhile, we can similarly compute the features for the P-frames $\{P^1, P^2, \cdots, P^T\}$ with a more lightweight model than the one used for I-frame, by directly taking the motion vectors $M^t$ and residuals $R^t$, \ie, ${\bf x}_{M}^t = f_M(M^t)$, and ${\bf x}_{R}^t=f_R(R^t)$ as input, where ${\bf x}_{M}^t, {\bf x}_{R}^t \in \mathbb{R}^{C \times H \times W}$ denote the features of the motion vectors and residuals, respectively. In this way, a considerable amount of time can be saved on extracting features for the P-frames. This simple strategy can only bring limited performance gain \cite{DBLP:conf/cvpr/coviar}. The method \cite{DBLP:conf/cvpr/dmc-net} attempts to integrate the optical flow in training phase, which can further improve the accuracy. However, there is still much room for improvement of the aforementioned methods. Specifically, the motion vectors record the motion patterns of both the scenes and objects in videos, and the residuals provide the compensation information. Both of them do not contain the context information of scenes. To that end, we design the spatial-channel compressed encoder module by integrating the features of the reference I-frame $x_I$ in computing the features of P-frames.

\begin{figure}[t]
  \centering
   \includegraphics[width=1.0\linewidth]{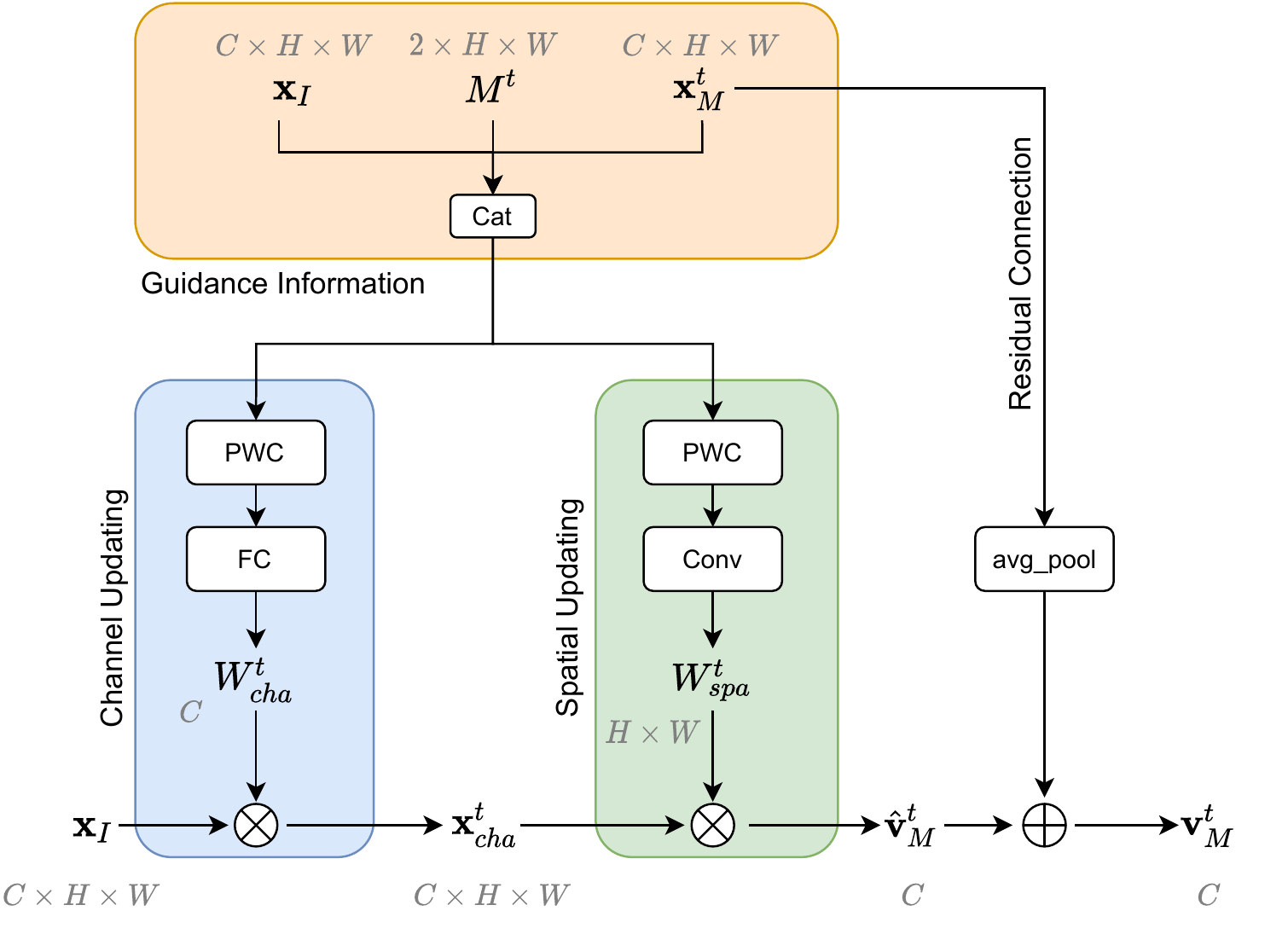}
   \caption{The architecture of the proposed spatial-channel compressed encoder (SCCE) module. We concatenate the features of I-frame ${\bf x}_I$, motion vectors $M^t$ and the features of motion vectors ${\bf x}_M^t$ to modulate the features of the reference I-frame ${\bf x}_I$ in both channel and spatial dimensions. After that, the modulated features $\hat{{\bf v}}_M^t$ is residually added with the features of motion vectors to obtain the refined vector representations ${\bf v}_M^t$.}
   \label{fig:scce}
   \vspace*{-0.15in}
\end{figure}

We first compute the features ${\bf x}_{M}^t$ of the motion vectors by refining the features of the reference I-frame ${\bf x}_I$ in both the channel and spatial dimensions. As indicated by \cite{DBLP:conf/eccv/ZeilerF14}, different regions on the feature maps focus on different parts of images. Thus, we introduce the attention weight for each feature map of ${\bf x}_I$ based on the information of the P-frame ${\bf x}_{M}^t$. Specifically, we concatenate I-frame feature ${\bf x}_I$, motion vector feature ${\bf x}_{M}^t$ and motion vectors $M^t$ together in the channel dimension to compute the channel weight $W_\text{cha}^t$ using a lightweight PWC-Net \cite{DBLP:conf/cvpr/SunY0K18}, \ie,
\begin{equation}
\begin{array}{ll}
    W_\text{cha}^t &= \sigma(W_2 \cdot \zeta(W_1h_\text{cha}^t+b_1)+b_2) \\
    {\bf h}_\text{cha}^t &= \operatorname{AvgPool}({\bf z}_\text{cha}^t) \\
    {\bf z}_\text{cha}^t &= \operatorname{PWC}([{\bf x}_I; {\bf x}_{M}^t; M^t])
\end{array}
\end{equation}
where $\sigma$ is the sigmoid function, $\zeta$ is the ReLU function, and $W_1, b_1, W_2, b_2$ are the learnable weights of the FC layers. After that, the features of the I-frame ${\bf x}_I$ are updated based on $W_\text{cha}^t$ as follows.
\begin{equation}
\begin{array}{ll}
    {\bf x}_\text{cha}^t = {\bf x}_I \otimes W_\text{cha}^t,
\end{array}
\end{equation}
where $\otimes$ is the channel-wise multiplication. In this way, we can compute the channel-weighted feature ${\bf x}_\text{cha}^t$ by updating ${\bf x}_I$ in channel dimension, with the guidance of the motion vectors. Meanwhile, the channel-weighted feature ${\bf x}_\text{cha}^t$ is further updated in the spatial dimension and the spatial dimension is reduced. That is, given the features ${\bf x}_I$ of the reference I-frame, motion vector features ${\bf x}_{M}^t$ and motion vectors $M^t$, we compute the 2D weight map $W_\text{spa}^t$, \ie, 
\begin{equation}
\begin{array}{ll}
    W_\text{spa}^t &= \operatorname{softmax}({\bf h}_\text{spa}^t) \\
    {\bf h}_\text{spa}^t &= \operatorname{2DConv}({\bf z}_\text{spa}^t) \\
    {\bf z}_\text{spa}^t &= \operatorname{PWC}([{\bf x}_I; {\bf x}_{M}^t; M^t])
\end{array}
\end{equation}
where $W_\text{spa}^t \in \mathbb{R}^{H \times W}$ is the spatial weight map. After that, we use $W_\text{spa}^t$ to weight the features ${\bf x}_\text{cha}^t$ in the spatial dimension to compute the enriched features of the motion vectors $\hat{{\bf v}}_M^t \in \mathbb{R}^C$, \ie, 
\begin{equation}
\begin{array}{ll}
    \hat{{\bf v}}_M^t = \sum_{p=1}^{H \cdot W}{{\bf x}_\text{cha}^t \cdot W_\text{spa}^t},
\end{array}
\end{equation}
where $p$ enumerates all spatial positions of ${\bf x}_\text{cha}^t \cdot W_\text{spa}^t$. Finally, we add $\hat{{\bf v}}_M^t$ to the original features ${\bf x}_M^t$ of the P-frame to obtain the refined features of the motion vectors ${\bf v}_M^t \in \mathbb{R}^C$, \ie,
\begin{equation}
\begin{array}{ll}
    {\bf v}_M^t = \hat{{\bf v}}_M^t + \operatorname{AvgPool}({\bf x}_M^t).
\end{array}
\end{equation}
The overall computing process of ${\bf v}_M^t$ is presented in Figure \ref{fig:scce}. Similarly, we can compute the refined features for the residuals ${\bf v}_R^t \in \mathbb{R}^C$. The final feature representations for the P-frame is further computed as
\begin{equation}
\begin{array}{ll}
    \widetilde{\bf v}^t = {\bf v}_M^t + {\bf v}_R^t.
\end{array}
\end{equation}
In this way, we can compute the features of the P-frames $\{\widetilde{\bf v}^1, \widetilde{\bf v}^2, \cdots, \widetilde{\bf v}^T\}$ in the GOP by considering the reference I-frame $I$ in both channel and spatial dimensions. The overall process is very efficient and can be processed in parallel in GOPs. After extracting the discriminative features for both the I-frames and P-frames in the same feature space, we can predict the event boundaries efficiently and accurately.

\subsection{Temporal Contrastive Module}
\label{sec:temporal_parser}
Based on the extracted features of the video ${\cal V}$, we design the temporal contrastive module to predict the event boundaries. Inspired by humans, \ie, look back and forth around the candidate boundary frames to determine event boundaries, we compute the contrastive features before and after the candidate boundary frames in the temporal domain. Specifically, given feature representations $\{\widetilde{\bf v}^{l-k}, \widetilde{\bf v}^{l-(k-1)}, \cdots, \widetilde{\bf v}^{l-1}\}$ before $k$ frames of candidate boundary frame $l$, we compute the left features $\phi^l$ of the candidate boundary frame $l$ using the simple linear weighted summation strategy, \ie,
\begin{equation}
\begin{array}{ll}
    \phi^l = \sum_{j=1}^{k}W_j \cdot \widetilde{\bf v}^{l-j},
\end{array}
\end{equation}
where $W_j \in \mathbb{R}^C$ is the learnable weights and shared at different position $l$. The simple linear weighted summation can be efficiently implemented using the 1D convolutional operation. Meanwhile, the right features $\psi^l$ can be similarly computed, \ie, weighted summing the feature representations of the $k$ features after the candidate boundary frame $l$. After that, the contrastive feature $\chi^l$ is computed as the concatenation of $\phi^l$ and $\psi^l$, \ie, $\chi^l = [\phi^l; \psi^l]$. Then for the final classification, we use the contrastive representations $\{\chi^1, \chi^2, \cdots, \chi^L\}$ to make the event boundary predictions.

\subsection{Loss Function}
Given the feature representations $\{\chi^1, \chi^2, \cdots, \chi^L\}$ of each video frame and the corresponding ground-truth labels, the event boundary detection task is intuitively formulated as the binary classification task. However, the ambiguities of annotations disrupt the learning process, which leads to poor convergence. To solve this issue, we use the Gaussian kernel to preprocess the ground-truth event boundaries to obtain the soft labels instead of using the ``hard labels'' of boundaries. Specifically, for each annotated boundary, the intermediate label of the neighboring position $i$ is computed as:
\begin{equation}
    g_i^l = \exp\Big( -\frac{( l-i )^2}{2\alpha^2} \Big)
\end{equation}
where $g_i^l$ indicates the intermediate label at time $i$ corresponding to the annotated boundaries at time $l$. We set $\alpha =1$ in all our experiments. The final soft labels are computed as the summation of all intermediate labels. Finally, a simple nonlinear Conv1D classifier is applied to predict the boundary score $S^l$ and the binary cross-entropy loss is used to guide the training process.

\begin{figure}[t]
  \centering
   \includegraphics[width=1.0\linewidth]{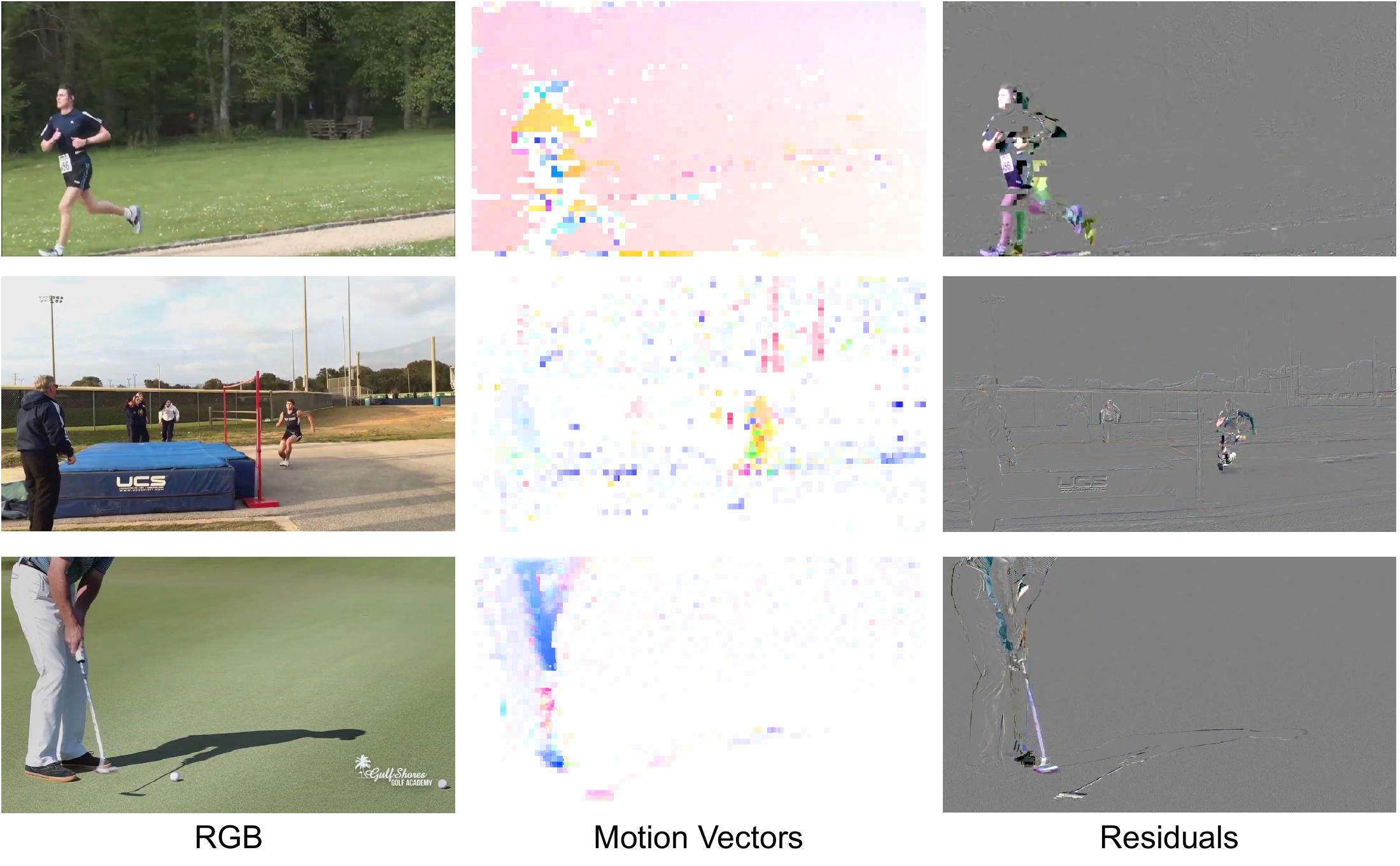}
   \caption{Visualization of the compressed information. The decoded RGB frames, motion vectors and residuals are presented in different columns. Best view in color.}
   \label{fig:compressed_visualization}
   \vspace*{-0.15in}
\end{figure}

%------------------------------------------------------------------------
\vspace*{-0.1in}
\section{Experiments}
\vspace*{-0.1in}
\label{sec:experiments}
{\noindent \textbf{Implementation detail.}} ResNet50 and ResNet18 \cite{DBLP:conf/cvpr/resnet} pretrained on ImageNet \cite{DBLP:conf/cvpr/imagenet} are used to extract the features for I-frames and P-frames in all experiments if not particularly indicated. Our method is implemented based on the MPEG-4 Part 2 specifications \cite{DBLP:journals/cacm/Gall91}, where each GOP contains $1$ I-frame and $11$ P-frames. We sample $3$ P-frames in each GOP to reduce the redundancy, \ie, $T=3$ in \eqref{equ:video_representation}. We use the standard SGD with momentum set to $0.9$, weight decay set to $10^{-4}$, and learning rate set to $10^{-2}$. We set the batch size to $4$ for each GPU and train the network on $8$ NVIDIA Tesla V100 GPUs, resulting in a total batch size of $32$. The network is trained for $30$ epochs with a learning rate drop by a factor of $10$ after $16$ epochs and $24$ epochs, respectively. We test the running speed of all methods on $1$ NVIDIA Tesla V100 GPU. All the source code of our method will be made publicly available after the paper is accepted.  

\begin{table}[t]
\caption{Accuracy on the HMDB-51 and UCF-101 datasets for both decoded video based methods and compressed video based methods. Our spatial-channel compressed encoder (SCCE) performs favorably against the state-of-the-art compressed video based methods.}
\centering
\setlength{\tabcolsep}{3pt}
\small {
% \normalsize
\begin{tabular}{lcc}
\hline
\multicolumn{1}{c}{} & \multicolumn{1}{c}{HMDB-51} & \multicolumn{1}{c}{UCF-101} \\ \hline
\multicolumn{3}{l}{\textbf{Decoded video based methods} \textbf{\textit{(RGB only)}}} \\
% \multicolumn{2}{c}{\textit{\textbf{Frame-level classification}}} & \\
ResNet-50 \cite{DBLP:conf/cvpr/resnet} & 48.9 & 82.3 \\
ResNet-152 \cite{DBLP:conf/cvpr/resnet} & 46.7 & 83.4 \\
% \multicolumn{2}{c}{\textit{\textbf{Motion representation learning}}} & \\
ActionFlowNet (2-frames) \cite{DBLP:conf/wacv/NgCND18} & 42.6 & 71.0 \\
ActionFlowNet \cite{DBLP:conf/wacv/NgCND18} & 56.4 & 83.9 \\ 
PWC-Net (ResNet-18) + CoViAR \cite{DBLP:conf/cvpr/SunY0K18} & 62.2 & 90.6 \\
TVNet \cite{DBLP:conf/cvpr/FanHGEGH18} & 71.0 & 94.5 \\
% \multicolumn{2}{c}{\textit{\textbf{Spatio-temporal modeling}}} & \\
C3D \cite{DBLP:conf/iccv/TranBFTP15} & 51.6 & 82.3 \\
Res3D \cite{DBLP:journals/corr/abs-1708-05038} & 54.9 & 85.8 \\
ARTNet \cite{DBLP:conf/cvpr/WangL0G18} & 70.9 & 94.3 \\ 
MF-Net \cite{DBLP:conf/eccv/ChenKLYF18} & 74.6 & 96.0 \\
S3D \cite{DBLP:journals/corr/abs-1712-04851} & 75.9 & 96.8 \\ 
I3D RGB \cite{DBLP:conf/cvpr/CarreiraZ17} & 74.8 & 95.6 \\
\hline
\multicolumn{3}{l}{\textbf{Compressed video based methods}} \\
EMV-CNN \cite{DBLP:conf/cvpr/ZhangWW0W16} & 51.2 (split1) & 86.4 \\
DTMV-CNN \cite{DBLP:journals/tip/ZhangWWQW18} & 55.3 & 87.5 \\
CoViAR \cite{DBLP:conf/cvpr/coviar} & 59.1 & 90.4 \\
DMC-Net(ResNet-18) \cite{DBLP:conf/cvpr/dmc-net} & 62.8 & 90.9 \\ 
DMC-Net(I3D) \cite{DBLP:conf/cvpr/dmc-net} & 71.8 & 92.3 \\
\textbf{Ours (ResNet-18)}  & \textbf{63.3} & \textbf{91.0} \\
\textbf{Ours (I3D)}  & \textbf{72.1} &\textbf{92.5} \\
\hline
\end{tabular}}
\label{table:results_of_action_recognition}
\vspace*{-0.15in}
\end{table}

{\noindent \textbf{Datasets.}} We conduct our experiments on the Kinetics-GEBD dataset \cite{DBLP:journals/corr/GEBD}, which contains the largest number of temporal boundaries. The Kinetics-GEBD dataset includes $54691$ videos and $1,290,000$ event boundaries, spans a broad spectrum of video domains in the wild and is open-vocabulary rather than building on a pre-defined taxonomy. Besides, to verify the generality and effectiveness of our method, we also conduct experiments on the popular action recognition datasets UCF101 \cite{DBLP:journals/corr/UCF101} and HMDB51 \cite{DBLP:conf/iccv/HMDB-dataset}. UCF101 consists of 101 action classes over $13,320$ videos and HMDB51 contains $51$ distinct action categories with a total of $6,766$ video clips.

\begin{table*}[t]
\caption{The evaluation results on the Kinetics-GEBD validation set with different Rel.Dis. threshold. Our method improves the F1 score over all thresholds by a large margin.}
\vspace*{-0.1in}
\centering
\setlength{\tabcolsep}{8.0pt}
\small{
\begin{tabular}{c|cccccccccc|c}
\hline
Rel.Dis. Threshold & 0.05  & 0.1  & 0.15  & 0.2 & 0.25 & 0.3  & 0.35 & 0.4 & 0.45 & 0.5 &avg \\
\hline
BMN\cite{DBLP:conf/iccv/LinLLDW19} & 0.186  & 0.204 & 0.213 & 0.220 & 0.226 & 0.230 & 0.233 & 0.237 & 0.239 & 0.241 &0.223 \\
BMN-StartEnd\cite{DBLP:journals/corr/GEBD} & 0.491 & 0.589 & 0.627 & 0.648 & 0.660 & 0.668 & 0.674 & 0.678 & 0.681 &0.683 &0.640 \\
TCN-TAPOS\cite{DBLP:journals/corr/GEBD} &0.464  &0.560  &0.602  &0.628  &0.645  &0.659  &0.669  &0.676  &0.682  &0.687 &0.627 \\
TCN\cite{DBLP:conf/eccv/LeaRVH16} & 0.588 & 0.657 & 0.679 & 0.691 & 0.698 & 0.703 & 0.706 & 0.708 & 0.710 &0.712 &0.685   \\
PC\cite{DBLP:journals/corr/GEBD}  & 0.625 &0.758 &0.804 &0.829 &0.844 &0.853 &0.859 &0.864 &0.867 &0.870 &0.817 \\
PC + Optical Flow &0.646 & 0.776 & 0.818 & 0.842 & 0.856 & 0.864 & 0.868 & 0.874 & 0.877 & 0.879 & 0.830 \\
\hline
Ours &\textbf{0.743}&\textbf{0.830} & \textbf{0.857} & \textbf{0.872} & \textbf{0.880} &\textbf{0.886}  & \textbf{0.890} & \textbf{0.893} & \textbf{0.896} &\textbf{0.898} & \textbf{0.865}\\
\hline
\end{tabular}}
\label{tab:results_of_kinetics_val}
\vspace*{-0.1in}
\end{table*}

\subsection{Discussion}
{\noindent \textbf{Kinetics-GEBD.}} We first train and evaluate the proposed method on the Kinetics-GEBD \cite{DBLP:journals/corr/GEBD} train-validation split. The evaluation protocol presented in \cite{DBLP:journals/corr/GEBD} uses Relative Distance (\ie, \textbf{Rel.Dis.}, the error between the predicted and ground truth timestamps) to determine whether a prediction is correct or not and then use the precision, recall, and F1 scores as the evaluation metrics. The results are shown in Table \ref{tab:results_of_kinetics_val}. Compared to the previous method PC \cite{DBLP:journals/corr/GEBD}, our method achieves 11.8\% absolute improvement while running $10\times$ faster. Meanwhile, we also add an additional optical flow input stream to PC. A slight improvement is observed after integrating optical flow, which indicates that the motion information (\ie, optical flow) alone can only provide limit temporal information for the generic event boundary detection task. Using motion vectors and residuals, our method provides more information features of the compressed P-frames by considering both the spatial and channel dimensions. The performance gap between the PC with optical flow and the proposed method demonstrates that the proposed method provides strong temporal cues for GEBD explicitly.

{\noindent \textbf{UCF101 and HMDB51.}} To validate the effectiveness of our method, we also conduct experiments on the action recognition datasets UCF-101 and HMDB-51. We follow the same settings as CoViAR \cite{DBLP:conf/cvpr/coviar} except that we use spatial-channel compressed encoder to process the motion vectors and residuals. Note that our temporal contrastive module is designed to capture temporal dependency, which is more suitable for event boundary detection. Thus, it is not applied in the action recognition task. The results are shown in Table \ref{table:results_of_action_recognition}. Our method achieves competitive results comparing with the state-of-the-art methods in compressed domain, \ie, EMV-CNN \cite{DBLP:conf/cvpr/ZhangWW0W16}, DTMV-CNN \cite{DBLP:journals/tip/ZhangWWQW18}, CoViAR \cite{DBLP:conf/cvpr/coviar} and DMC-Net \cite{DBLP:conf/cvpr/dmc-net}. We believe that this is because our method is able to generate more discriminative P-frame representations with the help of the proposed SCCE module. In contrast to other methods that process motion vectors and residuals in separate branches from the reference I-frame, we integrate the features of I-frame with the guidance of motion vectors and residuals on both spatial and channel dimensions. In this way, the rich information from the features of I-frame, motion vectors and residuals could be effectively fused together to generate high quality P-frame features with little overhead. It's worth noting that DMC-Net \cite{DBLP:conf/cvpr/dmc-net} needs extra optical flow as supervision during training phase while our method can directly learn discriminative features with the spatial-channel compressed encoder.

\subsection{Ablation Study}
We conduct several ablation studies to demonstrate the effectiveness of different components in the proposed method. All experiments are conducted on the Kinetics-GEBD train split with ResNet50 backbone and tested on a local minval split to reduce the computation cost. The local minval split is constructed from the Kinetics-GEBD validation split by randomly sampling $2000$ videos.

\begin{table}[ht]
\caption{The effectiveness of our proposed end-to-end architecture. ``E2E'' indicates end-to-end training strategy and ``GS'' indicates using soft labels generated by the Gaussian kernel strategy. To study the influence of these modules, we simply replace the inputs of the PC method by down-sampling each video into a succession of frames and use a Gaussian kernel to smooth the labels. This strategy improves the accuracy of the PC \cite{DBLP:journals/corr/GEBD} method with a much faster running speed by reducing redundant computations.}
\vspace*{-0.1in}
\centering
\setlength{\tabcolsep}{10.0pt}
\small {
\begin{tabular}{lcccc}
\hline
  &Rec &Prec &F1 &Speed(ms) \\
\hline
PC \cite{DBLP:journals/corr/GEBD} &0.611 &0.631 &0.621 &46.4 \\
+ E2E    &0.629 &0.640 &0.634 &9.3 \\
+ GS     &0.665 &0.643 &0.654 &9.3 \\
\hline
\end{tabular}}
\label{tab:ablation_end_to_end}
\vspace*{-0.15in}
\end{table}

{\noindent \textbf{The end-to-end architecture.}} The previous PC method \cite{DBLP:journals/corr/GEBD} formulate GEBD as the classification task, which feeds preceding and succeeding frames as inputs to provide temporal context information. To verify the effectiveness and the efficiency of the proposed end-to-end architecture, we conduct experiments with the identical architecture as PC \cite{DBLP:journals/corr/GEBD} except the feature inputs and target labels, shown in Table \ref{tab:ablation_end_to_end}. Simply replacing the feature inputs of PC \cite{DBLP:journals/corr/GEBD} with continuous video frames gives 1.3\% absolute performance gain while increasing inference speed by a large margin, which indicates that sharing features between nearby frames is beneficial for the GEBD task. Besides, using the soft labels generated by Gaussian kernel provides further 2.0\% absolute improvements. Using the ambiguous ``hard labels'' disrupt the learning process, which leads to poor convergence. Our soft label strategy effectively solve this issue and speeds up the training process.

\begin{table}[ht]
\caption{Ablation study of different compressed representation. ``OF'' indicates the optical flow and ``Vanilla'' indicates using the vanilla ResNet-18 to extract the features of motion vectors and residuals. We observe that both the methods improved from PC \cite{DBLP:journals/corr/GEBD} and our method benefit from optical flow and motion vectors and residuals. The proposed spatial-channel compressed encoder (SCCE) module further improve the accuracy with similar running speed.}
\centering
\setlength{\tabcolsep}{6.0pt}
\small {
\begin{tabular}{l|ccccc}
\hline
Method  &Repre. &Rec &Prec &F1 &Speed(ms) \\
\hline
\multirow{4}{*}{PC \cite{DBLP:journals/corr/GEBD}} &-    &0.611    &0.631  &0.621  & 46.4   \\
                          &OF   &0.635    &0.658  &0.646 & 69.3   \\
                          &Vanilla &0.643 &0.641  &0.642 & 33.2   \\
                          &SCCE    &0.709 &0.638  &0.669 & 34.5  \\
\hline
\multirow{4}{*}{Ours}     &-      &0.665     &0.643  &0.654  & 9.3  \\
                          &OF     &0.649     &0.673  &0.661  & 15.7  \\
                          &Vanilla  &0.659  &0.656  &0.657  & \textbf{4.1}  \\
                          &SCCE      &0.725 &0.651  &\textbf{0.686}  & 4.5 \\
\hline
\end{tabular}}
\label{tab:ablation_compressed}
\vspace*{-0.1in}
\end{table}

{\noindent \textbf{Compressed representation.}} We conduct the ablation studies on various strategies of using the compressed representations, \ie, (1) use optical flow (OF) in PC \cite{DBLP:journals/corr/GEBD}, (2) replace RGB images of P-frames in PC \cite{DBLP:journals/corr/GEBD} with compressed representation (\ie, motion vectors and residuals), (3) use our spatial-channel compressed encoder in PC \cite{DBLP:journals/corr/GEBD}, (4) remove compressed representation in our method, (5) use optical flow in our method, and (6) replace spatial-channel compressed encoder in our method with a vanilla encoder. The results are shown in Table \ref{tab:ablation_compressed}. Visualization exemplars of the compressed information are shown in Figure \ref{fig:compressed_visualization}. Both PC \cite{DBLP:journals/corr/GEBD} and our method are benefit from optical flow, compressed representation and the spatial-channel compressed encoder module respectively. Specifically, the optical flow branch improves F1 score by $2.5\%$ compared to the original PC \cite{DBLP:journals/corr/GEBD} method, with a much slower inference speed. The compressed information only brings limited improvements to PC \cite{DBLP:journals/corr/GEBD} and our end-to-end method with a relatively faster inference speed. This phenomenon indicates that the simple usage of motion vectors and residuals cannot fully exploit the rich information contained in compressed domain. The proposed spatial-channel compressed encoder provides significant performance improvements without extra computation cost, indicating that the proposed encoder can learn more discriminative features of P-frames. This module allows our proposed method to fully exploit the compressed representations and capture crucial motion information from the nearly cost-free motion vectors and residuals.

\begin{table}[t]
\caption{Ablation study of our temporal contrastive module by varying the window size $k$. $k=0$ means we remove the temporal contrastive module. This study shows that it's critical to learn the temporal dependency explicitly for event boundary detection. However, the value of window size gives limited influence to the performance.}
\vspace*{-0.1in}
\centering
\setlength{\tabcolsep}{10.0pt}
\small{
\begin{tabular}{l|ccccc}
\hline
Window size &Rec &Prec &F1 \\
\hline
 $k=0$  &0.725 &0.651 &0.686 \\
 $k=2$  &0.675 &0.749 &0.710 \\
 $k=4$  &0.697 &0.745 &0.720 \\
 $k=6$  &0.729 &0.744 &0.736 \\
 $k=8$  &\textbf{0.757} &0.736 &\textbf{0.746} \\
 $k=10$ &0.725 &0.750 &0.737 \\
$k=12$  &0.696 &\textbf{0.761} &0.727 \\
\hline
\end{tabular}}
\label{tab:ablation_temporal}
\vspace*{-0.15in}
\end{table}

{\noindent \textbf{Temporal contrastive module.}} Besides the discriminative features of P-frames, the temporal dependencies are also important to predict the accurate event boundaries. To validate the effectiveness of the temporal contrastive module, we conduct several experiments, shown in Table \ref{tab:ablation_temporal}. As shown in Table \ref{tab:ablation_temporal}, without the temporal contrastive module (\ie, $k=0$), the overall accuracy (F1 score) decreased dramatically. After adapting the proposed temporal module, the F1 score improves sharply, \ie, $0.710$ {\it vs.} $0.686$ at $k=2$. To further analyze the effective of different window size in model accuracy, we also perform several experiments with different $k$ values. Table \ref{tab:ablation_temporal} shows that the recall starts to drop when $k>8$. We believe that it is because larger window size mixes temporal information cross boundaries, resulting in the combination of multiple different predictions and decreasing the recall value. Considering the performance, we set $k=8$ in our experiments as the default setting.

\begin{table}[t]
\vspace*{-0.1in}
\caption{Comparisons with the state-of-the-art methods. The results are evaluated in validation split. $\dagger$ indicate the results come from our implementations since the test server is unavailable now. $\dagger$ CLA \cite{DBLP:journals/corr/abs-2106-11549} uses the concatenation of pre-trained two-stream TSN \cite{DBLP:conf/eccv/WangXW0LTG16} and
SlowFast \cite{DBLP:conf/iccv/Feichtenhofer0M19} features as input, $\dagger$ CASTANET \cite{DBLP:journals/corr/abs-2107-00239} and ours uses pretrained CSN \cite{DBLP:conf/iccv/TranWFT19} as backbone. The speed is computed by averaging per-frame decoding and inference time.}
\centering
\small{
\begin{tabular}{l|cccccc}
\hline
Method &Rec &Prec &F1 & Speed(ms)\\
\hline
$\dagger$ CLA \cite{DBLP:journals/corr/abs-2106-11549} &0.815 &0.768 &\textbf{0.791}& 90.2 \\
$\dagger$ CASTANET \cite{DBLP:journals/corr/abs-2107-00239} &0.838 &0.732 &0.781& 93.9 \\
\hline
 Ours (CSN+R18)  &0.813 &0.761 &0.786 &20.4 \\
 Ours (R50+R18)  &0.751 &0.742 &0.746 &\textbf{4.7} \\
\hline
\end{tabular}}
\label{tab:compare_sota}
\vspace*{-0.15in}
\end{table}

{\noindent \textbf{Comparisons with the state-of-the-arts.}} We compare the proposed method with the state-of-the-art methods at CVPR’21 \textbf{LO}ng-form \textbf{V}id\textbf{E}o \textbf{U}nderstanding (LOVEU) Challenge\footnote{https://sites.google.com/view/loveucvpr21}. CLA \cite{DBLP:journals/corr/abs-2106-11549} uses contrastive learning based approach to deal with the GEBD and utilizes temporal self-similarity matrix (TSM) as an intermediate representation. However, their approach relies on pre-extracted features and uses global similarity matrix, which hurts model's scalability. CASTANET \cite{DBLP:journals/corr/abs-2107-00239} adapts the identical framework from PC \cite{DBLP:journals/corr/GEBD} except the feature extractor and thus introduces redundant computations between nearby frames. Our method with ResNet50 runs extremely fast, \ie, $20\times$ faster than CLA \cite{DBLP:journals/corr/abs-2106-11549}. After replacing I-frame feature extractor $f_I$ with a more powerful backbone CSN \cite{DBLP:conf/iccv/TranWFT19}, we achieve competitive result, \ie, 0.787 compared with CLA 0.795 and CASTANET 0.784, while improving inference speed for more than $4\times$. The result shows the efficiency of working on the compressed domain using a lightweight network as the P-frame feature extractor and the effectiveness of our proposed method on high quality representation learning.

%------------------------------------------------------------------------
\vspace*{-0.05in}
\section{Conclusion}
\vspace*{-0.05in}
\label{sec:conclusion}
In this work, we propose an end-to-end compressed video representation learning method for GEBD. Specifically, we convert the video input into successive frames and use the Gaussian kernel to preprocess the annotations. Meanwhile, we design a spatial-channel compressed encoder to make full use of the motion vectors and residuals to learn discriminative feature representations for P-frames. After that, we propose a temporal contrastive module to model the temporal dependency between frames and generate accurate event boundaries. Extensive experiments have conducted on the Kinetics-GEBD dataset demonstrate that the proposed method performs favorably against the state-of-the-art methods.

\vspace*{-0.05in}
\section{Acknowledgement}
\vspace*{-0.05in}
This work was supported by the Key Research Program of Frontier Sciences, CAS, Grant No. ZDBS-LY-JSC038. Libo Zhang was supported CAAI-Huawei MindSpore Open Fund and Youth Innovation Promotion Association, CAS (2020111).

%%%%%%%% REFERENCES
{\small
% \bibliographystyle{ieee_fullname}
% \bibliography{references}

}

\end{document}